\pdfoutput=1

\documentclass[11pt]{article}

\usepackage[]{ACL2023}

\usepackage{times}
\usepackage{latexsym}

\usepackage[T1]{fontenc}

\usepackage[utf8]{inputenc}

\usepackage{microtype}

\usepackage{inconsolata}

\usepackage{enumitem}
\usepackage{refstyle}
\usepackage{xspace}
\usepackage{colortbl}

\usepackage{booktabs}
\usepackage{multirow} %
\usepackage{soul}%
\usepackage{tabularx}

\usepackage{amsmath}
\DeclareMathOperator*{\argmax}{argmax} %
\usepackage{pifont}

\usepackage{graphicx}

\usepackage{cleveref}
\crefformat{section}{\S#2#1#3}
\crefformat{subsection}{\S#2#1#3}
\crefformat{subsubsection}{\S#2#1#3}
\crefrangeformat{section}{\S#3#1#4 to~\S#5#2#6}
\crefmultiformat{section}{\S#2#1#3}{ and~\S#2#1#3}{, #2#1#3}{ and~#2#1#3}
\Crefformat{figure}{#2Fig.~#1#3}
\Crefmultiformat{figure}{Figs.~#2#1#3}{ and~#2#1#3}{, #2#1#3}{ and~#2#1#3}
\Crefformat{table}{#2Tab.~#1#3}
\Crefmultiformat{table}{Tabs.~#2#1#3}{ and~#2#1#3}{, #2#1#3}{ and~#2#1#3}
\Crefformat{appendix}{#2Appx.~\S#1#3}
\crefformat{algorithm}{Alg.~#2#1#3}
\Crefformat{equation}{#2Eq.~#1#3}

\newcommand{\stitle}[1]{\vspace{1ex} \noindent{\bf #1.}}

\newcommand{\Model}{\mbox{\textsc{Read}}\xspace}

\newcommand{\FigureExample}{
    \begin{figure}[t]
        \begin{center}
            \includegraphics[width=\columnwidth]{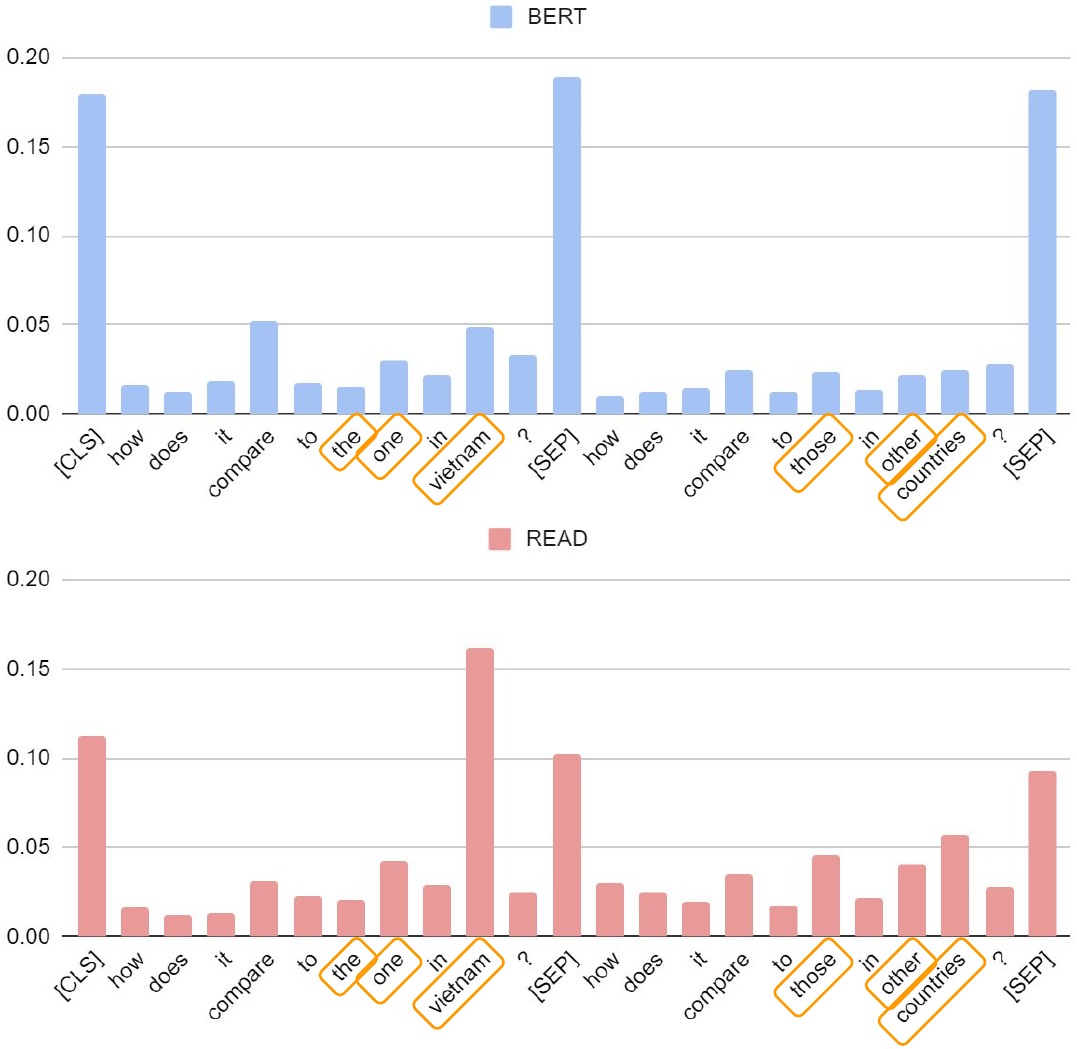} %
        \end{center}
        \caption{Attention distribution on a non-duplicated sentence pair. Red bars are debiased \texttt{[CLS]} attention from the last ensemble layer of \Model and blue bars are corresponding attention from finetuned BERT. Distinct tokens in the two sentences are highlighted with orange borderlines. \Model %
        pays more attention to distinct tokens and is more robust to lexical overlap bias.}
        \label{fig/example}
    \end{figure}
}

\newcommand{\FigureModel}{
    \begin{figure*}[t]
        \begin{center}
            \includegraphics[width=\textwidth]{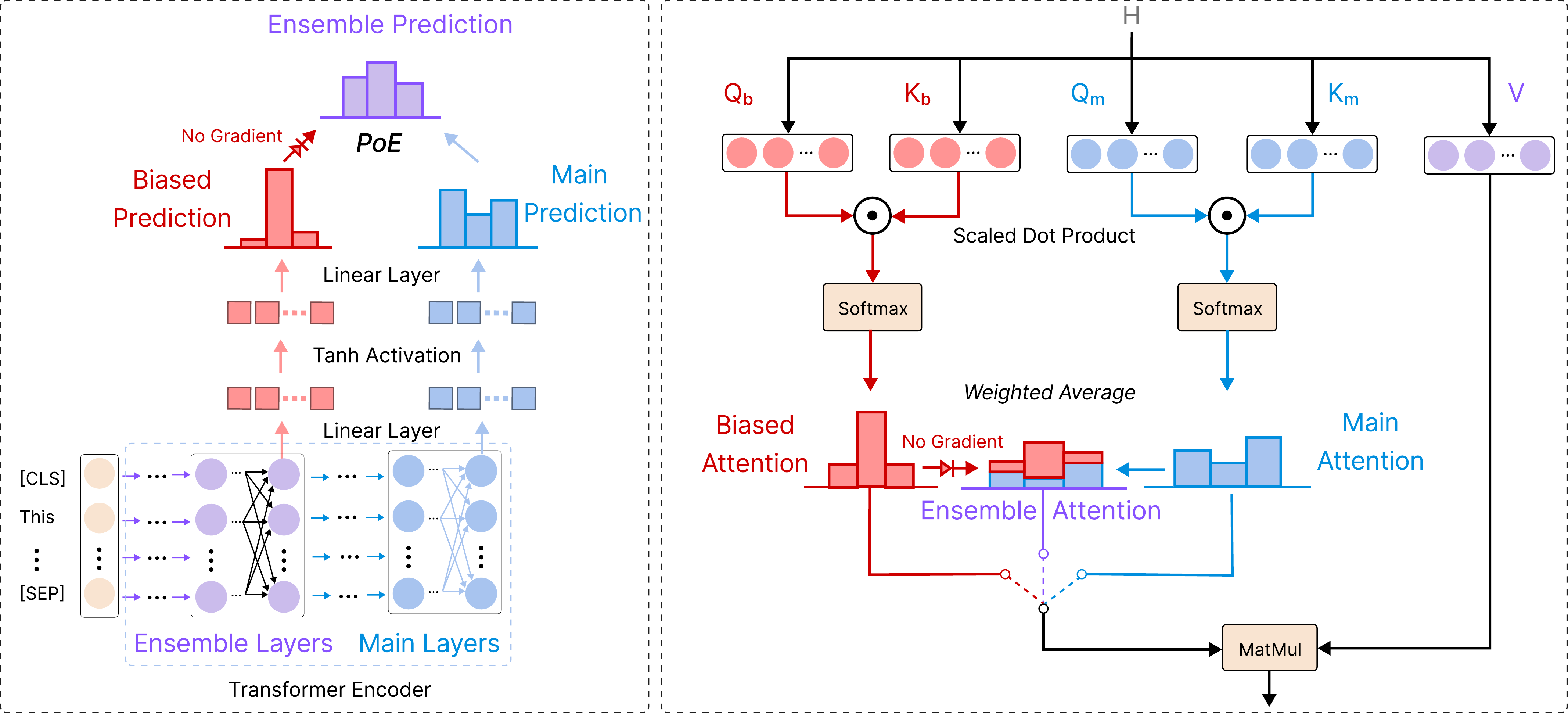} %
        \end{center}
        \caption{Illustration of one-stage PoE (left) and learning attention residual in ensemble layers (right). Dotted lines in the right figure are conditionally activated. During training, ensemble attention is activated to compute the main prediction and biased attention is activated to compute biased prediction. Through learning their residual, \Model mitigates biases from the main attention. During inference, the debiased main attention is activated to compute robust main prediction.}
        \label{fig/model}
    \end{figure*}
}

\newcommand{\TableMain}{
    \begin{table*}[t!]
        \setlength{\tabcolsep}{5pt}
        \centering
        \small
        \begin{tabular}{l|c>{\columncolor[RGB]{230, 242, 255}}c|c>{\columncolor[RGB]{230, 242, 255}}c|c>{\columncolor[RGB]{230, 242, 255}}c}
        \toprule
        \multirow{2}{*}{\textbf{Model}}  & \multicolumn{2}{c|}{\textbf{MNLI (Acc.)}} & \multicolumn{2}{c|}{\textbf{FEVER (Acc.)}}  & \multicolumn{2}{c}{\textbf{QQP (F1)}} \\
        & Dev & HANS & Dev & Sym. & Dev & PAWS \\
        \midrule
        BERT-base & 84.8$^\ddagger$ & 60.2$^\ddagger$ & 87.0$^\ddagger$ & 57.7$^\ddagger$ & 88.4$^\ddagger$ & 44.0$^\ddagger$ \\
        \midrule
        \multicolumn{7}{c}{\textit{Known Bias Mitigation}} \\
        \midrule
        Reweighting \cite{clark2019don} & 83.5 & 69.2 & -  & - & - & - \\
        PoE \cite{clark2019don} & 83.0 & 67.9 & - & - & -  & - \\
        DRiFt \cite{he2019unlearn} & 81.8$^\dagger$ & 66.5$^\dagger$ & 84.2$^\dagger$ & 62.3$^\dagger$ & - & - \\
        Conf-Reg \cite{utama2020mind} & 84.3 & 69.1 & 86.4 & 60.5  & - & 46.1$^*$ \\
        MoCaD \cite{xiong2021uncertainty} & 84.1 & 70.7 & 87.1 & 65.9 & - & - \\
        \midrule
        \multicolumn{7}{c}{\textit{Unknown Bias Mitigation}} \\
        \midrule
        PoE w/ Weak Learner  \cite{sanh2020learning} & 81.4 & 68.8$^*$ & 82.0 & 60.0 & - & - \\
        Self-Debias  \cite{utama2020towards} & 82.3 & 69.7 & - & - & - & - \\
        MoCaD  \cite{xiong2021uncertainty} & 82.3 & 70.7 & - & - & - & - \\
        End2End \cite{ghaddar2021end} & 83.2 & 71.2 & 86.9 & 63.8 & - & - \\
        Masked Debiasing \cite{meissner2022debiasing} & 82.2 & 67.9 & - & - & 89.6 & 44.3 \\
        DCT \cite{lyu2023feature} & 84.2 & 68.3 & 87.1 & 63.3 & - & - \\
        Kernel-Whitening \cite{gao2022kernel} & - & 70.9 & - & 66.2 & - & 45.2$^*$ \\
        \midrule
        \textbf{\Model} & 79.6 $\pm$ 0.7 & \textbf{73.1} $\pm$ 0.7 & 79.2 $\pm$ 1.9 & \textbf{68.7} $\pm$ 2.1 & 84.5 $\pm$ 0.3 & \textbf{46.7} $\pm$ 1.7 \\
        \Model ($p_e$) & 83.6 $\pm$ 0.3 & 64.8 $\pm$ 1.2 & 84.3 $\pm$ 1.1 & 55.3 $\pm$ 1.8 & 87.7 $\pm$ 0.0 & 44.8 $\pm$ 0.7 \\
        \bottomrule
        \end{tabular}
        \caption{Model performance on MNLI, FEVER, and QQP. We report results on both the in-distribution dev set and the OOD challenge set (highlighted in blue). All baseline results are copied from the referenced paper unless marked otherwise. For methods that have multiple variants, we report the variant with the best average OOD performance. $^\ddagger$ reproduced with our code base. $^*$ computed based on reported (subset) accuracy. $^\dagger$ copied from \citet{xiong2021uncertainty}.
        }
        \label{table/main} 
    \end{table*}
}

\newcommand{\FigureAtt}{
    \begin{figure*}[t]
        \begin{center}
           \includegraphics[width=0.32\textwidth, clip, trim=0.5cm 0.5cm 0.5cm 0.5cm]{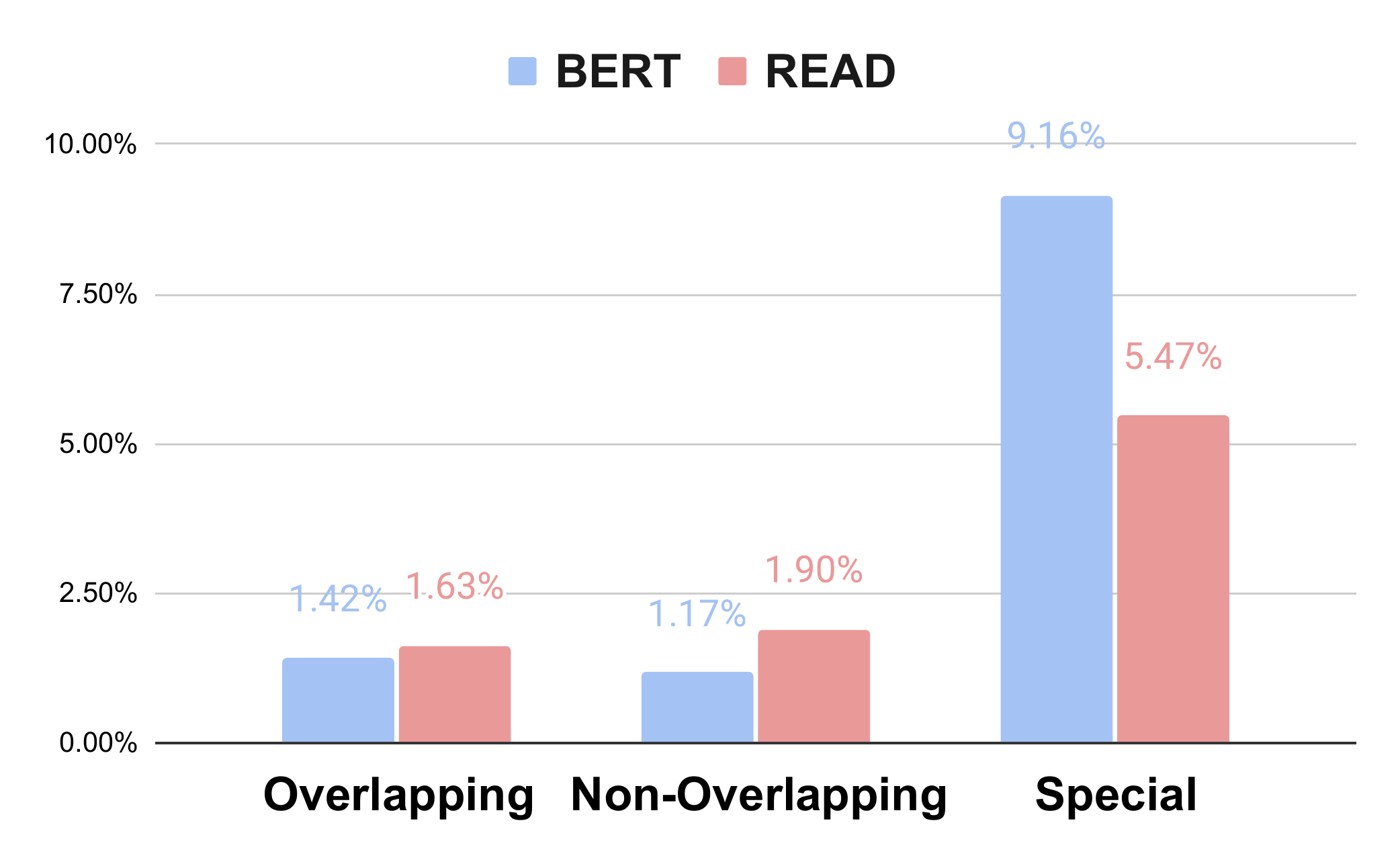} 
            \includegraphics[width=0.32\textwidth, clip, trim=0.5cm 0.5cm 0.5cm 0.5cm]{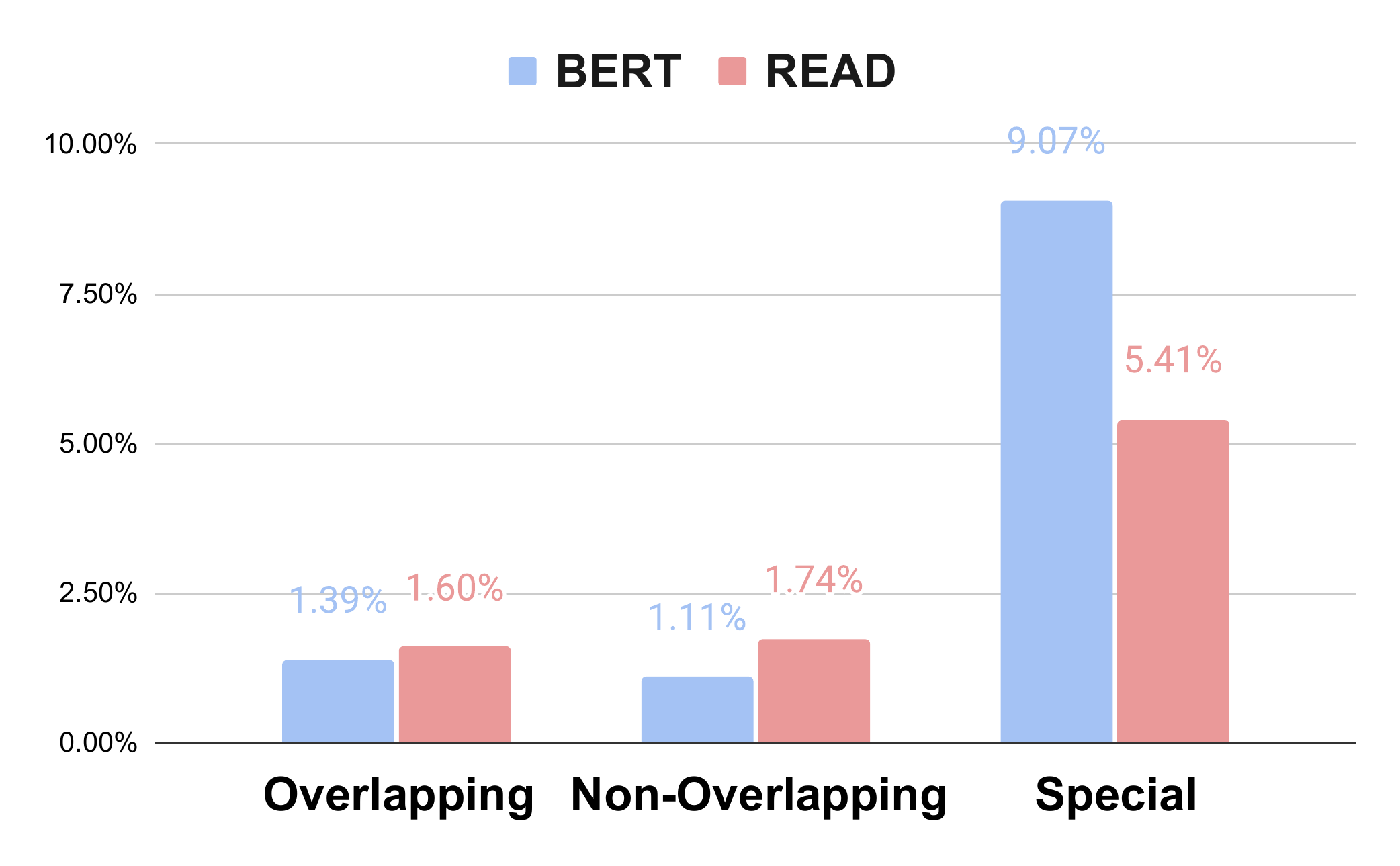} 
            \includegraphics[width=0.32\textwidth, clip, trim=0.5cm 0.5cm 0.5cm 0.5cm]{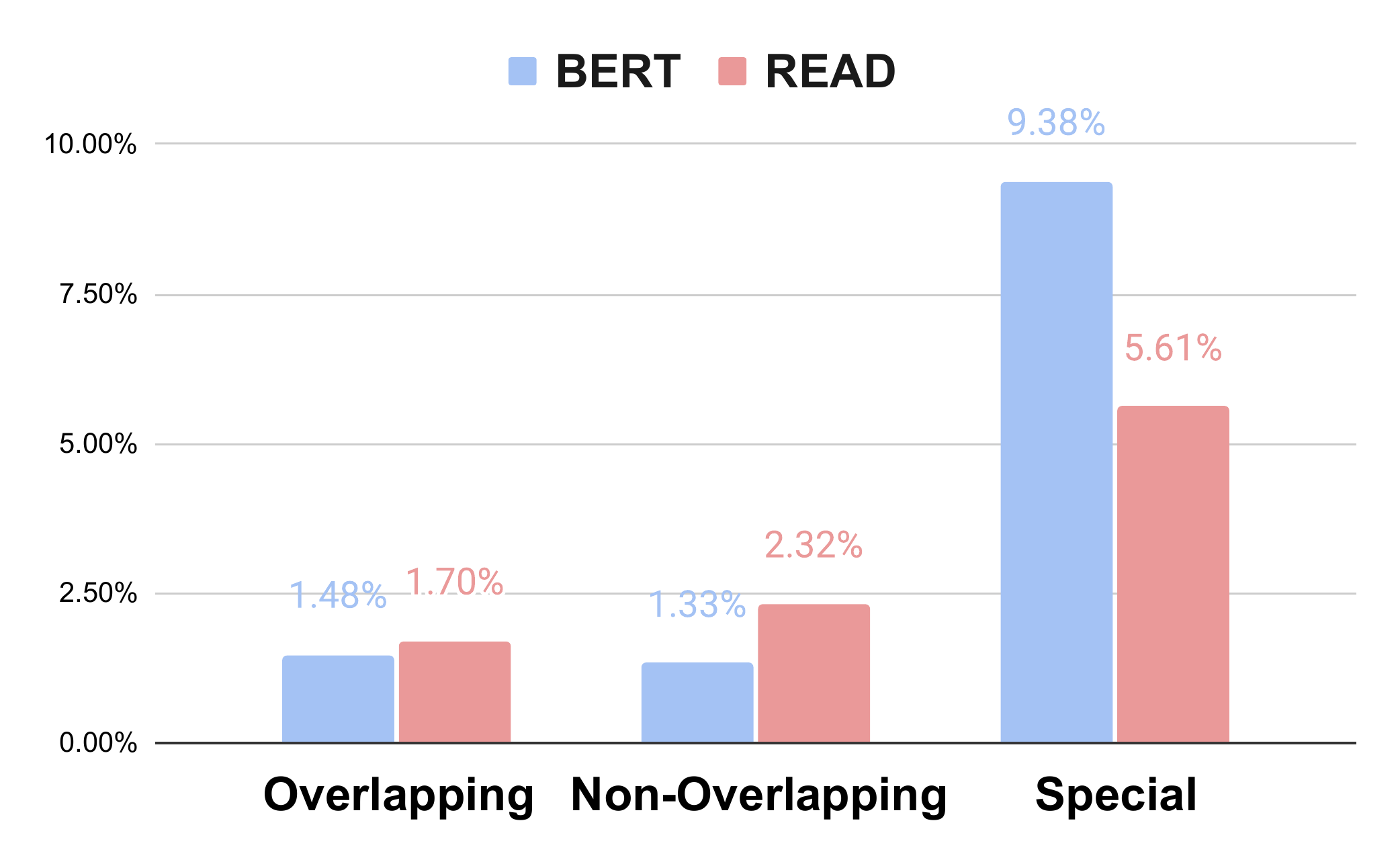} 
        \end{center}
        \caption{Average attention probability of each overlapping token, non-overlapping token, and special token per sentence pair on the PAWS test set for all instances (left), non-duplicated instances (middle), and duplicated instances (right). We present the \texttt{[CLS]} attention over all input tokens from the last ensemble layer of \Model and the same attention layer of BERT. \Model increases the attention over non-overlapping tokens to reduce lexical overlap bias.}
        \label{fig/att}
    \end{figure*}
}

\newcommand{\FigureLayer}{
    \begin{figure*}[t]
        \begin{center}
            \includegraphics[width=0.32\textwidth]{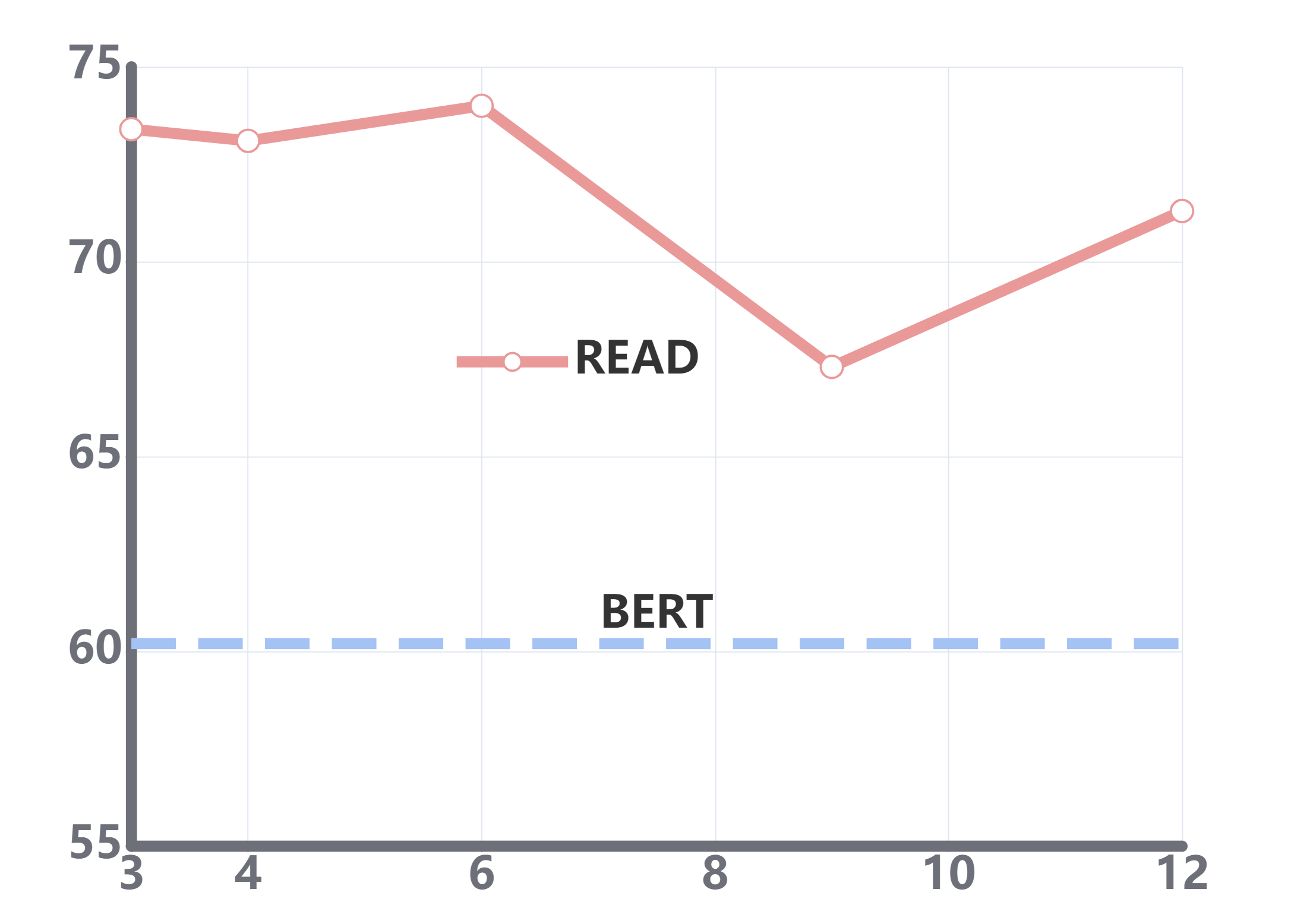} 
            \includegraphics[width=0.32\textwidth]{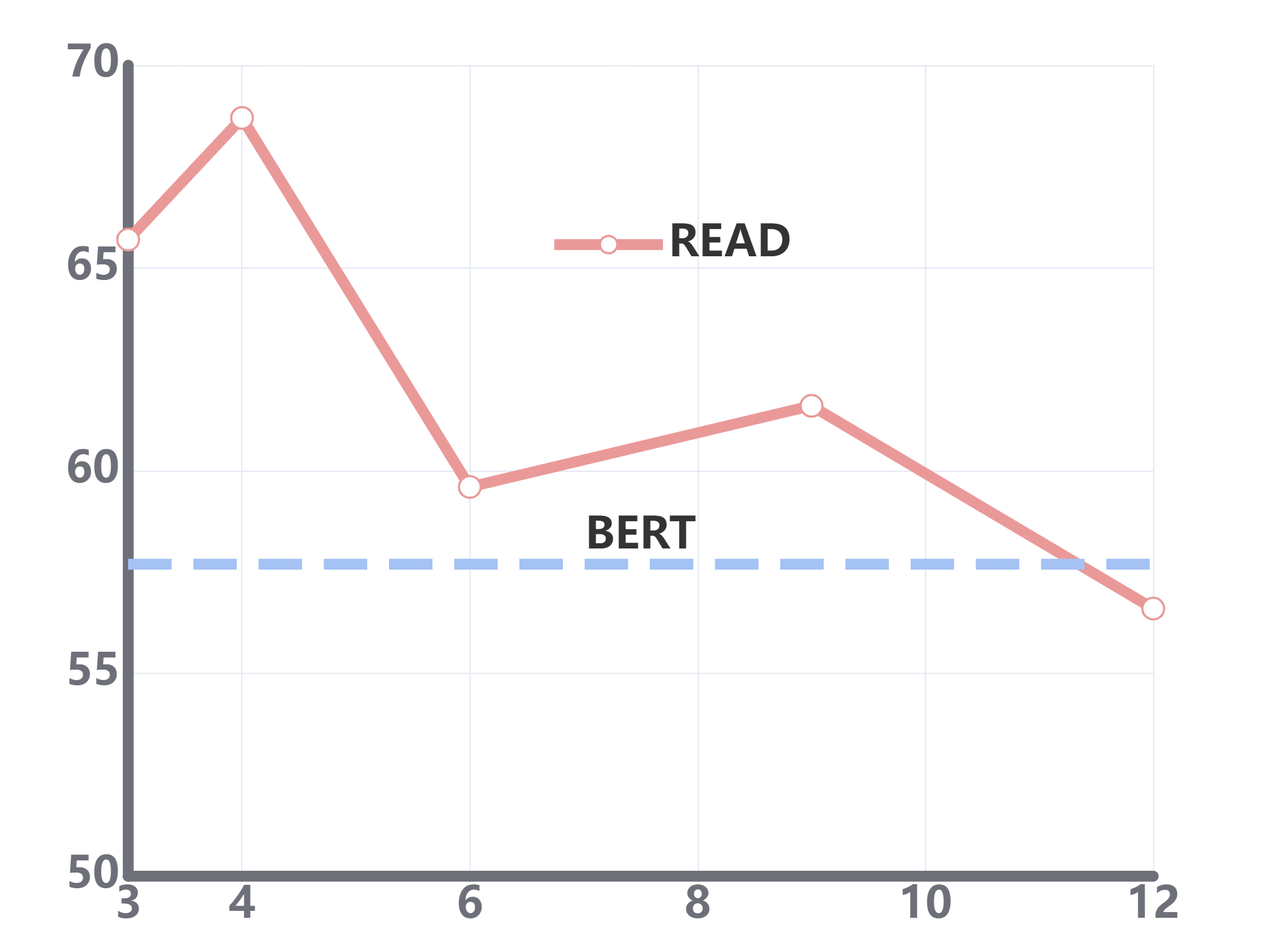} 
            \includegraphics[width=0.32\textwidth]{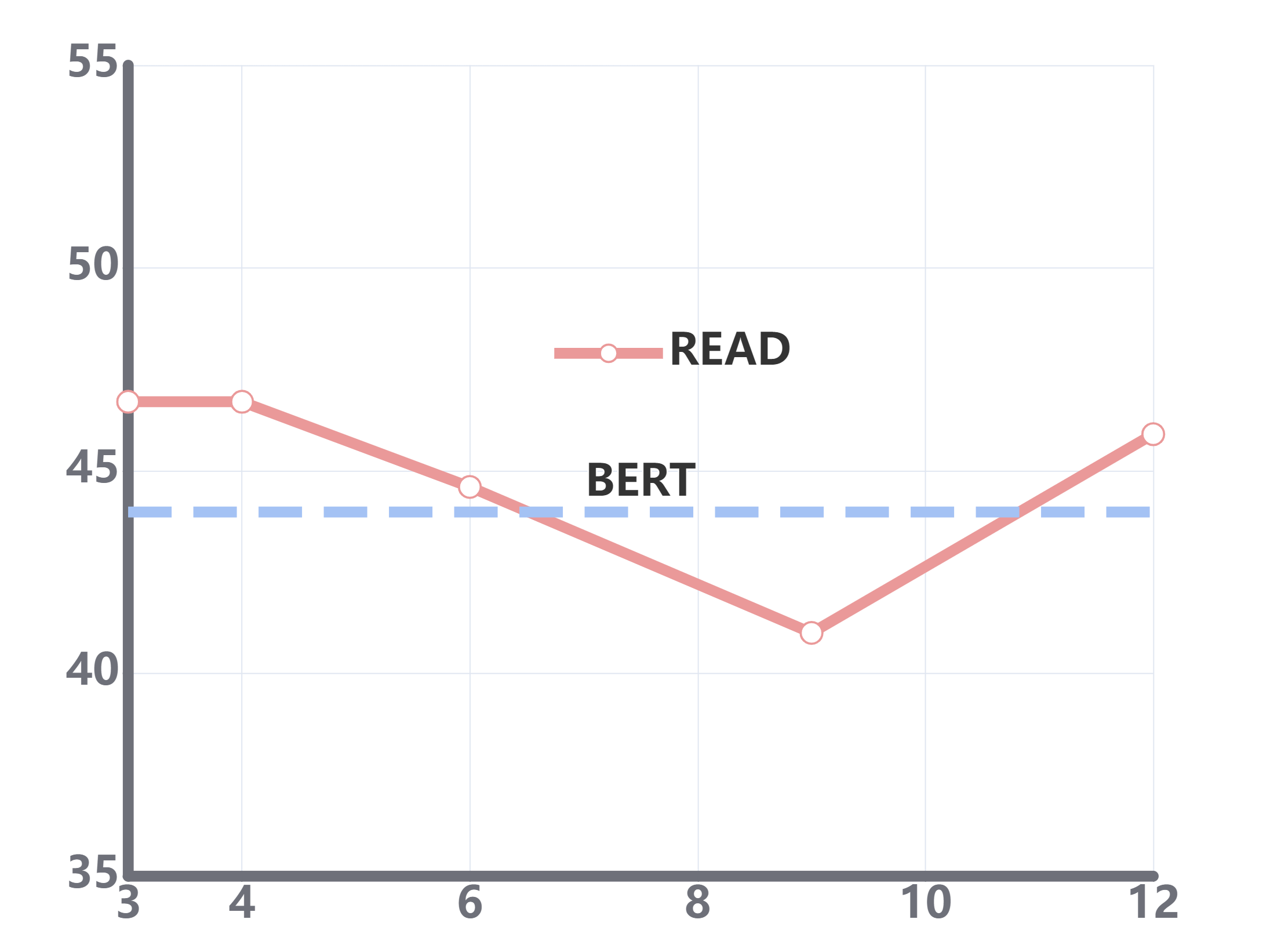} 
        \end{center}
        \caption{Performance of \Model by the number of ensemble layers on HANS (left), FEVER-Symmetric (middle), and PAWS (right). \Model is still effective when using twelve ensemble layers on HANS and PAWS.
        }
        \label{fig/layer}
    \end{figure*}
}

\title{\mbox{Robust Natural Language Understanding with Residual Attention Debiasing}}

\author{
  Fei Wang,$^*$ James Y. Huang,$^*$ Tianyi Yan, Wenxuan Zhou \and Muhao Chen
 \\
  University of Southern California\\
   \texttt{\{fwang598,huangjam,tianyiy,zhouwenx,muhaoche\}@usc.edu} \\ 
}

\begin{document}
\maketitle

\renewcommand{\thefootnote}{\fnsymbol{footnote}}
\footnotetext[1]{The first two authors contributed equally.}
\renewcommand{\thefootnote}{\arabic{footnote}}

\begin{abstract}

Natural language understanding (NLU) models often suffer from unintended dataset biases. Among bias mitigation methods, ensemble-based debiasing methods, especially product-of-experts (PoE), have stood out for their impressive empirical success. However, previous ensemble-based debiasing methods typically apply debiasing on top-level logits without directly addressing biased attention patterns. Attention serves as the main media of feature interaction and aggregation in PLMs and plays a crucial role in providing robust prediction. In this paper, we propose REsidual Attention Debiasing (\Model), an end-to-end debiasing method that mitigates unintended biases from attention. Experiments on three NLU tasks show that \Model significantly improves the performance of BERT-based models on OOD data with shortcuts removed, including +12.9\% accuracy on HANS, +11.0\% accuracy on FEVER-Symmetric, and +2.7\% F1 on PAWS. Detailed analyses demonstrate the crucial role of unbiased attention in robust NLU models and that \Model effectively mitigates biases in attention.\footnote{Code is available at \url{https://github.com/luka-group/READ}.}
    
\end{abstract}
\section{Introduction}

Natural language understanding (NLU) models often suffer from unintended dataset biases  \cite{jia2017adversarial,gururangan2018annotation,poliak2018hypothesis,gardner2021competency,rajaee2022looking},
causing them to learn spurious shortcuts and %
make unfaithful or under-generalized prediction
\cite{mccoy2019right,schuster2019towards,zhang2019paws}.
While a number of methods have been proposed to tackle this problem based on prior knowledge of specific biasing features \cite{clark2019don,he2019unlearn,mahabadi2020end,utama2020mind,liu2022win}, various unintended biases exist in NLU datasets, and not all of them are identifiable \cite{sanh2020learning,utama2020towards}. 
More recent works start to focus on mitigating unknown biases \cite{sanh2020learning,utama2020towards,xiong2021uncertainty,ghaddar2021end,meissner2022debiasing}. 
Among them, ensemble-based debiasing methods, especially product-of-experts (PoE), have stood out for their impressive empirical success \cite{sanh2020learning,utama2020towards,xiong2021uncertainty,ghaddar2021end}.

\FigureExample

Although the attention mechanism \cite{vaswani2017attention} is essential to the success of Transformer-based pretrained language models (PLMs), 
attention can also capture potentially spurious %
shortcut features leading to prediction biases.
For example, too much or too little attention across sentences in natural language inference may lead to the lexical overlap bias \cite{mccoy2019right,rajaee2022looking} or the hypothesis-only bias \cite{poliak2018hypothesis}.
Since attention serves as the main media for feature interactions in PLMs, many of the aforementioned biases can be associated with biased attention patterns.
In fact, a number of recent studies have shown that appropriate attention plays a critical role in ensuring robust\footnote{Robustness typically refers to the consistency of model behavior given original and (adversarially) perturbed inputs \cite{jia2019certified}, or given in-distribution and out-of-distribution (OOD) data \cite{hendrycks2020pretrained}. This paper focuses on OOD robustness, where OOD data do not share dataset biases with in-distribution data.} prediction \cite{chen2020universal,li2020neural,stacey2022supervising}.
However, existing ensemble-based debiasing methods typically apply debiasing on top-level logits \cite{clark2019don,he2019unlearn,sanh2020learning,utama2020towards,ghaddar2021end}. These methods do not proactively mitigate attention biases, but instead, rely on debiasing signals being propagated from final predictions to the attention modules in a top-down manner. Top-level logits are highly compressed and the propagation may suffer from information loss, thus providing limited debiasing signal to low-level attention. Instead, we seek for an effective attention debiasing method that prevents models from learning spurious shortcuts, especially those captured by the attention mechanism.

In this paper, we propose REsidual Attention Debiasing (\Model), an end-to-end debiasing method that mitigates unintended biases from attention. Our method is inspired by the recent success of one-stage PoE \cite{ghaddar2021end}. As an ensemble-based debiasing method, it trains a biased model to capture spurious in-distribution shortcuts and trains the ensemble of the biased model and a main model to prevent the main model from relying on spurious shortcuts. To do this end-to-end, one-stage PoE trains the biased model and its ensemble with the main model simultaneously in a weight-sharing manner. In \Model, we let the two models share all weights except attention modules and classification heads, allowing the main model to fit the unbiased attention residual with respect to the attention of the biased model. Intuitively, since they are trained on the same batch of data at each iteration, biased model attention and main model attention are likely to capture similar spurious features, making their residual free of such biases. \Cref{fig/example} presents an example of the attention change. Given a non-duplicate sentence pair, BERT, which suffers from lexical overlap bias, does not aggregate much information from non-overlapping tokens. In contrast, \Model learns to pay more attention to informative non-overlapping tokens. 

Experiments on three NLU tasks show that \Model significantly improves the performance of BERT-based models on OOD data where common types of shortcuts are removed, including +12.9\% accuracy on HANS, +11.0\% accuracy on FEVER-Symmetric, and +2.7\% F1 on PAWS.
We further examine the attention scores of the debiased main model and find that its distribution is more balanced (\Cref{sec/analysis/attention}).
These results indicate the crucial role of unbiased attention in robust NLU models.
We also demonstrate that our method is still effective when using a biased model with the same parameter size as the main model (\Cref{sec/analysis/layer}), which differs from the previous assumption that the biased model of unknown biases should be weaker\footnote{Under-trained or under-parameterized.} \cite{sanh2020learning,utama2020towards}.

Our contributions are three-fold.
First, we propose \Model, an ensemble-based debiasing method for NLU models, mitigating attention biases through learning attention residual. 
Second, experiments on three NLU tasks consistently demonstrate that \Model can significantly improve the OOD performance of different NLU tasks with various dataset biases.
Third, detailed analyses provide useful insights for developing robust NLP models, including the importance and properties of unbiased attention, and the design of biased models in ensemble-based debiasing methods.

\section{Method}

\FigureModel

Our method, \Model, combines one-stage product-of-experts (PoE) with learning attention residual to mitigate unknown dataset biases for NLU tasks.
Based on the problem definition, we introduce the two key components of our method, followed by the details of training and inference.

\subsection{Problem Definition}
For a discriminative task,
given the dataset $D=\{x_i,y_i\}$, where $x_i$ is the raw input and $y_i$ is the gold label, our goal is to learn a robust function $f$ with parameters $\theta$, that can predict a probability distribution $\mathbf{p}=f(x_i;\theta)$ without relying on spurious features. 
in NLU tasks, $x_i$ is typically a textual sequence. 
As discussed by prior studies \cite{gardner2021competency,eisenstein2022informativeness},
spurious features captured by $f$, such as particular words \cite{gardner2021competency} and lexical overlap ratios \cite{rajaee2022looking}, although may be statistically correlated with $y_i$ due to dataset artifacts \cite{gururangan2018annotation},
should not be regarded as useful information for predicting $y_i$. 
In other words, the prediction of a robust and faithful model should be independent of these non-causal features. Since diverse spurious features may exist in NLU datasets, we focus on mitigating dataset biases without any assumption of the type or structure of bias, so that the proposed method is generalizable to most unknown biases.

\subsection{One-Stage PoE} 
Due to the automated process of feature extraction in neural networks, it is impractical to train a robust model that directly identifies all robust features in the tremendous feature space.
Considering spurious features are often simple features (also, of easy-to-learn data instances) that model tends to memorize in the first place \cite{shah2020pitfalls}, an ensemble-based debiasing method trains a biased model to collect biased prediction $p_b$ %
and approximates an ensemble prediction $p_e$ based on $p_b$ and another prediction $p_m$ from a main model towards the observations in training data.
Considering both parts of the ensemble prediction $p_e$, since the biased model mainly captures spurious shortcuts, as its complement, the main model then focuses on capturing robust features.
\Model adopts PoE \cite{clark2019don} to obtain a multiplicative ensemble of the two models' predictions:
\begin{equation}
\label{eq/pe}
p_e \propto p_b p_m.
\end{equation}
Specifically, \Model follows the one-stage PoE framework \cite{ghaddar2021end} that simultaneously optimizes the ensemble prediction and biased prediction, and shares weights between the main model and biased model, as shown in \Cref{fig/model}.
When using a PLM as the main model, one-stage PoE typically uses one or a few bottom layers of PLMs stacked with an independent classification head as the biased model, because these low-level layers preserve rich surface features \cite{jawahar2019does} which can easily cause unintended biases \cite{mccoy2019right,gardner2021competency}.
The main model has shared encoder layers at the bottom followed by independent encoder layers and its classification head.
This weight-sharing design makes it possible to debias the model end-to-end with a few additional parameters. 
However, shared layers result in shared biases in these layers. Although PoE mitigates biases from predictions, it preserves biases in shared layers.

\subsection{Learning Attention Residual}
Ensemble prediction with PoE cannot effectively mitigate unintended biases in attention, which is the major part of feature aggregation and interaction in PLMs.
For example, the \texttt{[CLS]} representation aggregates information from all token representations according to the attention distribution, and all token representations interact with each other based on the attention values. 
Therefore, biased attention becomes the direct source of many spurious features, such as lexical overlap in natural language inference and semantic-neutral phrases in sentiment analysis \cite{friedman2022finding}.
To prevent the main model from learning biased attention, \Model further conducts additive ensemble of the attention distributions of both the main and biased models. 
Similar to ensemble prediction, the attention ensemble 
here encourages the main model attention to learn from the residual of biased model attention, %
so as to mitigate the biases captured by the latter from the former.

\Cref{fig/model} shows the workflow of learning attention residual. 
The self-attention mechanism \cite{vaswani2017attention} allows each vector in a matrix to interact with all vectors in the same matrix.
Specifically, the input matrix $H$ is first projected to a query matrix $Q$, a key matrix $K$, and a value matrix $V$.
Attention scores of all vectors to each vector is a probability distribution computed based on the dot product between $Q$ and $K$.
With attention scores as weights, the self-attention module maps each vector in $H$ to the weighted average of $V$.
In \Model, the main attention and biased attention use distinct projection weights for $Q$ and $K$, but take the same $H$ as inputs and share the same projection weights for $V$.
Distinct $Q$ and $K$ allow the two models to have their own attention.
Sharing $H$ and $V$ ensures the attention in the biased and main models are distributed in the same semantic space so that they are additive.\footnote{In contrast, if we use a completely independent attention module in the biased model, its attention will not be aligned with the main model attention.}

The ensemble attention $\mathbf{a_e}$ combines main attention $\mathbf{a_m}$ and biased attention $\mathbf{a_b}$ with weighted average.\footnote{Multiplicative ensemble (e.g. PoE), although works well on ensemble prediction, is unstable during training for attention ensemble and causes models to fail according to our observation. This phenomenon is related to the plausibility problem in \citet{li2022contrastive}. The fluctuation of a tiny probability on an uninformative token (e.g. a stop word) may significantly influence the result of PoE. Assuming we have simple distributions over two candidates $p_e = [10^{-8}, 1-10^{-8}]$ and $p_b = [10^{-6}, 1-10^{-6}]$, then according to \Cref{eq/pe}, the learned $p_m \approx [0.99, 0.01]$. Due to the probability change from $10^{-8}$ to $10^{-6}$ of the first candidate, the division between $p_e$ and $p_b$ maps the probability from extremely high (i.e. close to 1) to low (i.e. 0.01) and vice versa, i.e. over-debiasing. Such behavior is harmful to the learning process.}
This additive ensemble is inspired by the success of using the probability difference for post-hoc debiasing \cite{niu2021counterfactual,qian2021counterfactual,wang2022should} and preventing over-confidence \cite{miao2021prevent}.
In our case, the main attention is the difference between ensemble attention and biased attention.
\Model also adds a coefficient $\alpha \in (0,1)$ to balance the ensemble ratio. 
An appropriate coefficient can prevent over- or under-debiasing.
Finally, the ensemble attention becomes
\begin{equation}
\label{eq/ae}
\mathbf{a_e} = (1 - \alpha) \mathbf{a_m} + \alpha \mathbf{a_b}.
\end{equation}

Now that we have three paths in the attention module, including ensemble attention, main attention, and biased attention.
In each forward pass from the input to $p_m$ or $p_b$, only one of them is activated as the final attention distribution.
During training, \Model adopts ensemble attention to compute $p_m$ and biased attention to compute $p_b$, for mitigating biases from main attention by learning their residual.
During inference, \Model adopts main attention, which is free of bias, to compute robust prediction $p_m$.

\subsection{Training and Inference}
We train the ensemble model and the biased model on the same dataset batch $B$ simultaneously with a cross-entropy loss
\begin{equation}
\begin{split}
\mathcal{L} & = \mathcal{L}_e + \mathcal{L}_b \\
& = - \frac{1}{|B|} \sum_{i=1}^{|B|} \log p_{e}(y_i|x_i) + \log p_b(y_i|x_i).
\end{split}
\end{equation}
When minimizing $\mathcal{L}_e$, gradients on  $p_b$ in \Cref{eq/pe} and gradients on $\mathbf{a_b}$ in \Cref{eq/ae} are disabled, because they serve as auxiliary values %
for computing $p_e$.
Backward passes on $p_b$ and $\mathbf{a_b}$ are only allowed when minimizing $\mathcal{L}_b$.
During inference, only the main model is used to predict a label $\hat{y}_i$ from the label set $\mathcal{C}$:
\begin{equation}
\hat{y}_i = \argmax_{c=1}^{|\mathcal{C}|} p_m(c|x_i).
\end{equation}
\section{Experiment}
In this section, we evaluate the debiasing performance of \Model on three NLU tasks. We first provide an overview of the experimental settings (\Cref{Experiment/Datasets,Experiment/Implementation}), followed by a brief description of baseline methods (\Cref{Experiment/Baseline}). Finally, we present a detailed analysis of empirical results (\Cref{Experiment/Results}).

\subsection{Datasets} \label{Experiment/Datasets}
Following previous studies \cite{utama2020towards,ghaddar2021end,gao2022kernel}, we use three English NLU tasks for evaluation, namely natural language inference, fact verification, and paraphrase identification. %
Specifically, each of the tasks uses an out-of-distribution (OOD) test set where common types of shortcuts in the training data have been removed, in order to test the robustness of the debiased model.
More details can be found in \Cref{section/appendix/dataset}.

\textbf{MNLI} (Multi-Genre Natural Language Inference; \citet{williams2018broad})  is a natural language inference dataset. %
The dataset contains 392k pairs of premises and hypotheses for training, which are annotated with textual entailment information (\textit{entailment}, \textit{neutral}, \textit{contradiction}). For evaluation, we report accuracy on the MNLI dev set and the OOD challenge set HANS \cite{mccoy2019right}. HANS contains premise-hypothesis pairs that have significant lexical overlap, and therefore models with lexical overlap bias would perform close to an entailment-only baseline.

\textbf{FEVER} \cite{Thorne18Fever} is a fact verification dataset that contains 311k pairs of claims and evidence labeled with the validity of the claim with the given evidence as context. For OOD testing, we report accuracy on the FEVER-Symmetric\footnote{Version 1.} test set \cite{schuster2019towards} %
where each claim is paired with both positive and negative evidences to avoid claim-only bias\footnote{Models overly relying on misleading cues from the claims while ignoring evidence.} .

\textbf{QQP} is a paraphrase identification dataset
consisting of pairs of questions that are labeled as either \textit{duplicated} or \textit{non-duplicate} depending on whether one sentence is a paraphrased version of the other. For testing, we report F1 score on PAWS \cite{paws2019naacl}, which represents a more challenging test set containing non-duplicate question pairs with high lexical overlap. %

\TableMain

\subsection{Implementation} \label{Experiment/Implementation}
Following previous works \cite{utama2020towards,ghaddar2021end,gao2022kernel}, we use BERT-base-uncased model \cite{devlin2019bert} as the backbone of the debiasing framework. All experiments are conducted on a single NVIDIA RTX A5000 GPU. We use the same set of hyperparameters across all three tasks, with the learning rate, batch size, and ensemble ratio ($\alpha$) set to 2e-5, 32, and 0.1 respectively. We train all models for 5 epochs and pick the best checkpoint based on the main model performance on the in-distribution dev set. On each dataset, we report average results and standard deviations of five runs. More details can be found in \Cref{section/appendix/implementation}.

\subsection{Baseline} \label{Experiment/Baseline}
We include a vanilla BERT model and compare our method with a wide selection of previous debiasing methods for language models as follows: %
\begin{itemize}[leftmargin=*] %
\setlength\itemsep{-0.1em}
 \item \textit{Reweighting} \cite{clark2019don} first trains a biased model to identify biased instances. During main model training, the biased instances are down-weighted, which encourages the model to focus more on unbiased instances.
 \item \textit{PoE} \cite{clark2019don} and \textit{DRiFt} \cite{he2019unlearn} %
 both train an ensemble of the biased and main models to learn the unbiased residual logits. The biased model is trained on data observed with a specific type of bias. Unlike our proposed \Model, these methods do not directly address biased attention patterns.
 \item \textit{Conf-Reg} \cite{utama2020mind} applies logit smoothing to a biased model to improve distillation. It prevents the model from making overly-confident predictions that are likely biased.
 \item \textit{MoCaD} \cite{xiong2021uncertainty} applies model calibration to improve the uncertainty estimations of a biased model. This method is generally complementary to a variety of ensemble-based methods.
 \item \textit{PoE w/ Weak Learner} \cite{sanh2020learning} and \textit{Self-Debias} \cite{utama2020towards} propose to use under-parameterized and under-trained models as biased models for ensemble-based debiasing methods, such as PoE. Since these weak models tend to rely on spurious shortcuts in datasets, they are effective in mitigating unknown bias.   %
 \item \textit{End2End} \cite{ghaddar2021end} is an ensemble-based debiasing method that shares the bottom layers of the main model as the whole encoder of the biased model. It reweights instances based on model predictions and regularizes intermediate representations by adding noise. %
 \item \textit{Masked Debiasing} \cite{meissner2022debiasing} searches and removes biased model parameters that contribute to biased model predictions, leading to a debiased subnetwork.
 \item \textit{DCT} \cite{lyu2023feature} reduces biased latent features through contrastive learning with a specifically designed sampling strategy. %
 \item \textit{Kernel-Whitening} \cite{gao2022kernel} transforms sentence representations into isotropic distribution with kernel approximation to eliminate nonlinear correlations between spurious features and model predictions. %
\end{itemize}
In addition, previous methods can also be categorized based on whether prior knowledge of specific biased features, such as hypothesis-only and lexical overlap biases in NLI, is incorporated in the debiasing process. We accordingly group the compared methods when reporting the results (\Cref{table/main}) in the following two categories:
\begin{itemize}[leftmargin=*]
\setlength\itemsep{-0.1em}
\item Methods for \textit{known bias mitigation} have access to the biased features before debiasing and therefore can train a biased model that only takes known biased features as inputs. While each of the OOD test sets we use for evaluation is crafted to target one specific form of bias, biased features can be highly complex and implicit in real-world scenarios, which limits the applicability of these methods.
\item Methods for \textit{unknown bias mitigation} do not assume the form of bias in the dataset to be given. Our proposed method belongs to this category.
\end{itemize}

\FigureAtt

\subsection{Results} \label{Experiment/Results}
As shown in \Cref{table/main},
among all baselines, unknown bias mitigation methods can achieve comparable or better performance than those for mitigating known biases on OOD test sets of NLI and fact verification. 
Although all baseline methods improve OOD performance in comparison with vanilla BERT, there is not a single baseline method that outperforms others on all three tasks.

Overall, our proposed method, \Model, significantly improves model robustness and outperforms baseline methods on all OOD test sets with different biases. 
On HANS, the challenging test set for MNLI, our method achieves an accuracy score of 73.1\%, i.e. a 12.9\% of absolute improvement from vanilla BERT and a 1.9\% improvement from the best-performing baseline \textit{End2End}. Compared to \textit{End2End}, residual debiasing on attention of \Model directly debiases on the interactions of token-level features, leading to more effective mitigation of lexical overlap biases. 
On FEVER-Symmetric, \Model outperforms vanilla BERT by 11.0\% accuracy and outperforms the best-performing method \textit{Kernel-Whitening} by 2.5\%.
On PAWS, the challenging test set for paraphrase identification, \Model improves model performance by 2.7\% F1, and outperforms the best-performing baseline method \textit{Conf-Reg}, which relies on extra training data with lexical overlap bias. 
These results demonstrate the generalizability of \Model for mitigating various biases in different NLU tasks.

We also observe that the in-distribution performance of \Model is generally lower than baseline methods. In fact, almost all debiasing methods shown in \Cref{table/main} enhance OOD generalization performance at the cost of decreased in-distribution performance %
This aligns with the inherent trade-off between in-distribution performance and OOD robustness as shown by recent studies \cite{tsipras2018robustness,zhang2019theoretically}. The optimal in-distribution classifier and robust classifier rely on fundamentally different features, so not surprisingly, more robust classifiers with less distribution-dependent features perform worse on in-distribution dev sets. 
However, note that generalizability is even more critical to a learning-based system in real-world application scenarios where it often sees way more diverse OOD inputs than it uses in in-distribution training.
Our method emphasizes the effectiveness and generalizability of debiasing on unknown OOD test sets and demonstrates the importance of learning unbiased attention patterns across different tasks. In the case where in-distribution performance is prioritized, the ensemble prediction $p_e$ can always be used in place of the debiased main prediction $p_m$ without requiring any additional training. Future work may also explore to further balance the trade-off between in-distribution and OOD performance \cite{raghunathan2020understanding,nam2020learning,liu2021just}.
It is also worth noting that our method only introduces a very small amount of additional parameters, thanks to the majority of shared parameters between biased and main models.

\FigureLayer

\section{Analysis}
To provide a comprehensive understanding of key techniques in \Model, we further 
analyze the debiased attention distribution (\Cref{sec/analysis/attention}) 
and the effect of number of ensemble layers (\Cref{sec/analysis/layer}).

\subsection{Debiased Attention Distribution}
\label{sec/analysis/attention}
To understand the influence of \Model on attention, we examine the attention distribution of BERT and \Model on the PAWS test set.
Specifically, we take the attention between \texttt{[CLS]}, which serves as feature aggregation, and all other tokens as an example. 
We group tokens into three categories, including overlapping tokens (e.g. \texttt{how} and \texttt{does} in \Cref{fig/example}),  non-overlapping tokens (e.g. \texttt{one} and \texttt{those} in \Cref{fig/example}), and special tokens (e.g. \texttt{[CLS]} and  \texttt{[SEP]}).
Since attention residual for attention debiasing exists in ensemble layers of \Model, we compare the attention on the last ensemble layer of \Model and the corresponding layer of BERT.

As discussed in \Cref{Experiment/Results}, vanilla BERT finetuned on QQP suffers from the lexical overlap bias and does not generalize well on PAWS.
This problem is reflected in the inner attention patterns.
As shown in \Cref{fig/att}, BERT assigns less (-0.25\%) attention to non-overlapping tokens than to overlapping tokens on average.
In contrast, \Model increases the attention on non-overlapping tokens to larger than (+0.27\%) the attention on overlapping tokens.
The same observation also appears in the subset of duplicate sentence pairs and the subset of non-duplicate sentence pairs. 
This change in attention patterns reveals the inner behavior of \Model for effectively preventing the model from overly relying on the lexical overlap feature.

\subsection{Effect of Number of Ensemble Layers}
\label{sec/analysis/layer}
Some previous studies assume that the biased model in PoE for unknown bias should be weaker (i.e. less trained or less parameterized) than the main model so as to focus on learning spurious features \cite{sanh2020learning,utama2020towards}.
One-stage PoE follows this assumption, using the bottom layers of the main model as the encoder of the biased model \cite{ghaddar2021end}.
Since biased attention patterns may appear in any layer, including top layers, we examine whether this assumption holds for \Model.
Specifically, we evaluate \Model with different numbers of ensemble layers on three OOD evaluation sets. 

As shown in \Cref{fig/layer}, although the best-performing \Model variant has few ensemble layers, the configuration where the biased and main models share all encoder layers is still effective on HANS and PAWS. 
For example, on HANS, \Model achieves comparable performance with the previous state-of-the-art method when the biased and main models share all encoder layers.
This observation indicates that the shared encoder layer with distinct attention allows the biased model to focus on spurious attention patterns.
Moreover, it is apart from the assumption that a biased model is necessarily a weak model, such as the bottom layers of the main model with a simple classification head. 
Future works on ensemble-based debiasing can explore a larger model space for the biased model.

\section{Related Work}
We present two lines of relevant research topics, each of which has a large body of work, so we can only provide a highly selected summary.

\stitle{Debiasing NLU Models}
Unintended dataset biases hinder the generalizability and reliability of NLU models \cite{mccoy2019right,schuster2019towards,zhang2019paws}.
While a wide range of methods have been proposed to tackle this problem, such as knowledge distillation \cite{utama2020mind,du2021towards}, neural network pruning \cite{meissner2022debiasing,liu2022win}, and counterfactual inference \cite{udomcharoenchaikit2022mitigating},
ensemble-based methods \cite{clark2019don,he2019unlearn,lyu2023feature} stand out for their impressive empirical success.
Recent works extend ensemble-based methods, such as PoE, to mitigate unknown biases by training a weak model to proactively capture the underlying data bias, then learn the residue between the captured biases and original task observations for debiasing \cite{sanh2020learning,utama2020towards,ghaddar2021end}.
\citet{xiong2021uncertainty} further improves the performance of these methods using a biased model with uncertainty calibration. %
Nevertheless, most prior works only mitigate unintended biases from top-level logits, ignoring biases in low-level attention.

\stitle{Attention Intervention}
In current language modeling technologies, the attention mechanism is widely used to characterize the focus, interactions and aggregations on features \cite{bahdanau2015neural,vaswani2017attention}.
Although the interpretation of attention is under discussion \cite{li2016understanding,jain2019attention,wiegreffe2019attention}, it still provides useful clues about the internal behavior of deep, especially Transformer-based, language models \cite{clark2019does}.
Through attention intervention, which seeks to re-parameterize the original attention to represent a conditioned or restricted structure,
a number of works have successfully improved various model capabilities, such as long sequences understanding \cite{beltagy2020longformer,shi2021sparsebert,ma2022mega}, contextualizing entity representation \cite{yamada2020luke}, information retrieval \cite{jiang2022retrieval}, and salient content selection \cite{hsu2018unified,wang2022salience}. %
Some recent works also add attention constraints to improve model robustness towards specific distribution shifts, including identity biases \citep{pruthi2020learning,attanasio2022entropy,gaci2022debiasing} and structural perturbations \cite{wang2022robust}.

\section{Conclusion}

In this paper, we propose \Model, an end-to-end debiasing method that mitigates unintended feature biases through learning the attention residual of two models.
Evaluation on OOD test sets of three NLU tasks demonstrates its effectiveness of unknown bias mitigation and reveals the crucial role of attention in robust NLU models. 
Future work can apply attention debiasing to mitigate dataset biases in generative tasks and multi-modality tasks, such as societal biases in language generation \cite{sheng2021societal} and language bias in visual question answering \cite{niu2021counterfactual}.

\section*{Acknowledgement}

We appreciate the reviewers for their insightful
comments and suggestions.
This work was partially supported by the NSF Grant IIS 2105329, by the Air Force Research Laboratory under
agreement number FA8750-20-2-10002, by an Amazon Research Award and a Cisco Research Award.
Fei Wang was supported by the Annenberg Fellowship at USC.
Tianyi Yan was supported by the Center for Undergraduate Research in Viterbi Engineering (CURVE) Fellowship.
Muhao Chen was also supported by a subaward of the INFER Program through UMD ARLIS.
Computing of this work was partly supported by a subaward of NSF Cloudbank 1925001 through UCSD.
\section*{Limitation}

Although our experiments follow the setting of previous works, 
the experimented tasks, types of biases, languages, and backbone PLMs can be further increased. 
As we do not enforce additional constraints when learning attention residual, there is a potential risk of over-debiasing, which is currently controlled by ensemble ratio $\alpha$.
We implement the idea of residual attention debiasing based on the one-stage PoE framework because it is one of the most successful end-to-end debiasing methods for NLU models. However, the effectiveness of attention debiasing may not be limited to the specific debiasing framework.
Since the proposed method focuses on mitigating attention biases, it cannot be directly applied to PLMs without attention modules, such as BiLSTM-based PLMs \cite{peters-etal-2018-deep}. 
Moreover, the proposed debiasing method may also be effective to generative PLMs, such as T5 \cite{raffel2020exploring} and GPT-3 \cite{brown2020language}. We leave this for future work.

\bibliography{reference}
\bibliographystyle{acl_natbib}

\clearpage

\appendix

\section{Datasets}
\label{section/appendix/dataset}

We use all the datasets in their intended ways. 

\textbf{MNLI} dataset contains different subsets released under the OANC’s license, Creative Commons Share-Alike 3.0 Unported License, and  Creative Commons Attribution 3.0 Unported License, respectively. Among all the data entries, 392,702 samples are used for training. 9,815 and 9,832 samples from validation matched and validation mismatched subsets of MNLI respectively are used for evaluation. 

\textbf{HANS} is released under MIT License. The validation subset of HANS contains 30,000 data entries, which are used for OOD evaluation of natural language inference. 

\textbf{FEVER} follows the Wikipedia Copyright Policy, and Creative Commons Attribution-ShareAlike License 3.0 if the former is unavailable. 311,431 examples from the FEVER dataset are used to train the model.

\textbf{FEVER-Symmetric} test set with 717 samples is used as the OOD challenge set for fact verification.

\textbf{QQP}\footnote{The dataset is available at \url{https://quoradata.quora.com/First-Quora-Dataset-Release-Question-Pairs}} consists of 363,846 samples for training, and 40430 samples for in-distribution evaluation. 

\textbf{PAWS} dataset with 677 entries is used for OOD evaluation of paraphrase identification.

\section{Implementation}
\label{section/appendix/implementation}

Our Implementation is based on HuggingFace's Transformers \cite{wolf-etal-2020-transformers} and PyTorch \cite{paszke2019pytorch}. 
Since the training sets of three tasks are of roughly the same size, it takes about 5 to 6 hours to finetune the BERT-base model, which has around 110 million parameters, on each task.
Our ensemble model adds 5.3M parameters, a 4.8\% increase from the BERT-base model. These additional parameters will be removed after the completion of training.
During training, we use a linear learning rate scheduler and the AdamW optimizer \cite{loshchilov2018decoupled}. Models finetuned on the MNLI dataset will predict three labels, including \textit{entailment}, \textit{neutral}, and \textit{contradiction}. During inference on the OOD test set, we map the latter two labels to the \textit{non-entailment} label in HANS. 

\end{document}